\documentclass[sigconf]{acmart}

\copyrightyear{2026}
\acmYear{2026}
\setcopyright{cc}
\setcctype{by}
\acmConference[MM '26] {Proceedings of the 34th ACM International Conference on Multimedia}{November 10--14, 2026}{Rio de Janeiro, Brazil.}
\acmBooktitle{Proceedings of the 34th ACM International Conference on Multimedia (MM '26), November 10--14, 2026, Rio de Janeiro, Brazil}
\acmISBN{979-8-4007-2213-4/2026/11}
\acmDOI{10.1145/3767308.3835504}

\usepackage{algorithm}
\usepackage{algorithmic}
\usepackage{enumitem}
\usepackage{multirow}
\usepackage{makecell}
\usepackage{wrapfig}
\usepackage{colortbl}

\newcommand{\XC}[1]{}
\newcommand{\JY}[1]{}

\definecolor{mygray}{RGB}{230,230,230}

\begin{document}

\title{EchoCache: Energy-Guided Cross-Modal Caching for Efficient
Audio-Driven Video Generation}

\author{Jiayu Chen}
\orcid{0009-0002-5428-8647}
\affiliation{%
  \institution{Peking University}
  \city{Beijing}
  \country{China}}
\email{jiayu.chen.25@stu.pku.edu.cn}

\author{Xiaoyu Wu}
\orcid{0009-0000-5594-7187}
\affiliation{%
  \institution{Peking University}
  \city{Beijing}
  \country{China}}
\email{2400013181@stu.pku.edu.cn}

\author{Rongshan Gao}
\orcid{0009-0009-9696-6419}
\affiliation{%
  \institution{Taiyuan University of Technology}
  \city{Taiyuan}
  \country{China}}
\email{2023002338@link.tyut.edu.cn}

\author{Maoliang Li}
\orcid{0009-0008-6773-5067}
\affiliation{%
  \institution{Peking University}
  \city{Beijing}
  \country{China}}
\email{maoliang.li@stu.pku.edu.cn}

\author{Zihao Zheng}
\orcid{0009-0008-4624-2853}
\affiliation{%
  \institution{Peking University}
  \city{Beijing}
  \country{China}}
\email{zhengzihao@stu.pku.edu.cn}

\author{Xinhao Sun}
\orcid{0009-0008-9086-8714}
\affiliation{%
  \institution{Peking University}
  \city{Beijing}
  \country{China}}
\email{sunxinhao5513@gmail.com}

\author{Hailong Zou}
\orcid{0009-0004-0090-9553}
\affiliation{%
  \institution{Peking University}
  \city{Beijing}
  \country{China}}
\email{zouhailong26@stu.pku.edu.cn}

\author{Guojie Luo}
\orcid{0000-0003-4932-3655}
\affiliation{%
  \institution{Peking University}
  \city{Beijing}
  \country{China}}
\email{gluo@pku.edu.cn}

\author{Xiang Chen}
\authornote{Corresponding author.}
\orcid{0000-0003-2790-976X}
\affiliation{%
  \institution{Peking University}
  \city{Beijing}
  \country{China}}
\email{xiang.chen@pku.edu.cn}

% 该文件应包含完整的：
% \begin{abstract} ... \end{abstract}
\begin{abstract}
Audio-driven video generation (A2V) has achieved promising progress in synthesizing temporally coherent and audio-visually aligned videos, yet its inference remains expensive due to the iterative denoising process of diffusion models. 
    Existing caching methods mainly exploit temporal redundancy in visual features, while overlooking the cross-modal alignment of A2V, where audio drives visual generation with highly non-uniform temporal importance. 
In this paper, we identify two levels of misalignments in existing A2V caching methods: temporal-semantic and computation-storage. 
    To address them, we propose EchoCache, an energy-guided cross-modal caching framework for efficient A2V generation. 
    EchoCache leverages audio time-frequency energy as a saliency anchor to guide latent-level cache updates, and further introduces a dynamic timestep-latent caching mechanism with quantized cache management for joint efficiency and memory optimization. 
Extensive experiments on mainstream A2V models show that EchoCache consistently improves the latency-quality trade-off while preserving generation quality and audio-visual consistency. In particular, on Wan2.2-S2V over the EMTD benchmark, EchoCache achieves 2.46$\times$ speedup with the best overall performance. code is available at \url{https://github.com/IF-LAB-PKU/EchoCache}.
\end{abstract}



% 该文件放置 ACM CCS 工具生成的完整 CCSXML 和 \ccsdesc。
% \input{_txt/0_ccs}


\maketitle

% 作者较多时，缩短页眉作者列表
\renewcommand{\shortauthors}{Jiayu Chen et al.}

%=======================================%
%               Sections                %
%=======================================%

\begin{figure}
  \centering
  \includegraphics[width=3.3in]{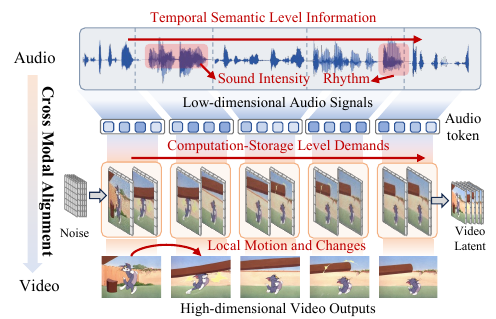}
\caption{Overview of Audio-driven Video Generation.}
  \label{fig:1}
  \vspace{-12pt}
\end{figure}

\section{Introduction}

\begin{figure*}
  \centering
  \includegraphics[width=\linewidth]{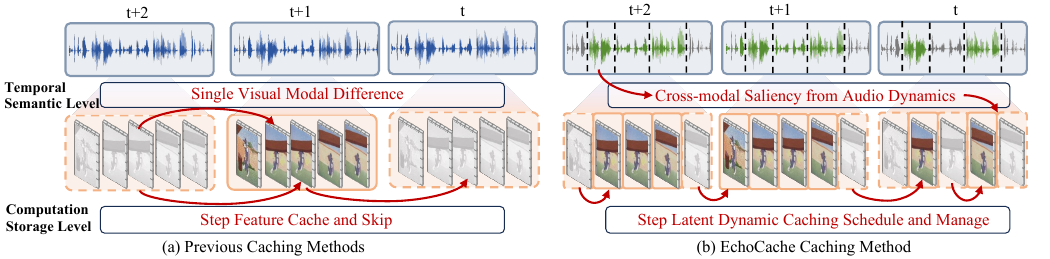}
\caption{Comparison of previous caching methods and the proposed EchoCache for audio-driven video generation.}
  \label{fig:2}
\end{figure*}

In recent years, DiT-based video generation models \cite{wan2025wan,zheng2024open,yang2024cogvideox} have leveraged their superior generation quality and spatiotemporal modeling capabilities to become an important paradigm in the video generation field.
    Among these models, audio-driven video generation~(A2V) models \cite{gao2025wan,meituanlongcatteam2025longcatvideoavatartechnicalreport,tian2025emo2} synthesize temporally coherent and audio-visually aligned videos based on input audio, demonstrating important application value in scenarios such as digital humans \cite{guan2025audcast,chen2025midas}, virtual avatars \cite{yu2025llia}, and cinematic content creation \cite{shen2024audioscenic}.
        Unlike text- or image-driven video generation, A2V is inherently a \textbf{cross-modal alignment} process in which low-dimensional audio signals drive the generation of high-dimensional visual content.
        The input audio provides not only global semantic guidance but also fine-grained temporal cues, such as speaking rate and rhythm, which govern the timing and intensity of local motions and changes in the generated video, as shown in Fig.~\ref{fig:1}.
However, this cross-modal upscaling process remains constrained by the inherently iterative and progressively refined nature of diffusion sampling.
    The model requires tens to hundreds of denoising steps over the full video latent sequence, resulting in substantial inference latency that hinders real-time A2V applications.

To alleviate this issue, caching has emerged as a promising plug-and-play approach for accelerating video generation.
    As shown in Fig.~\ref{fig:2}(a), existing methods reduce redundant computation by reusing intermediate visual features across adjacent sampling steps, using either static scheduling~\cite{li2023faster,ma2024deepcache,wimbauer2024cache} with fixed reuse intervals or dynamic scheduling~\cite{kahatapitiya2025adaptive,lv2024fastercache,liu2025speca} that adapts reuse decisions based on the difference between current visual and cached features.
Despite these advances, most existing methods assume temporal similarity within the visual modality and design scheduling solely around visual features, without fully exploiting the distinctive cross-modal interaction in A2V.
    In the A2V generation process, the model continuously performs cross-modal alignment between audio signals and visual synthesis.
    Audio affects video latents highly non-uniformly at both the \textit{temporal-semantic} and \textit{computation-storage} levels, leading to diverse update demands across temporal clips.

Based on this observation, we find that existing dynamic caching methods suffer from two levels of misalignment in the A2V model.
\textbf{(i) Temporal-semantic level}: a unified step-level coarse-grained strategy implicitly assumes that different latents within the current step require similar updates, while neglecting the differences in semantic importance and motion sensitivity among audio-driven temporal clips.
    As a result, dynamic and smooth clips are often treated uniformly, causing truly critical clips that carry local motion changes to be insufficiently updated.
(ii) \textbf{Computation-storage level}: cache scheduling based on visual feature differences ignores the temporal-semantic priors embedded in the audio modality, making it difficult to identify critical clips while storing non-critical ones, thereby limiting further efficiency gains.

Motivated by these observations, we argue that a more effective A2V acceleration strategy should return to the cross-modal nature of the task itself.
    Caching is no longer a passive reuse mechanism for visual redundancy, but rather an active computation allocation mechanism driven by cross-modal alignment.

To this end, we propose \textbf{EchoCache}, an \emph{energy-guided cross-modal caching framework} for efficient A2V generation, as shown in Fig.~\ref{fig:2}(b).
    The core idea of EchoCache is to explicitly transform the fine-grained temporal dynamics embedded in the audio modality into cross-modal alignment signals that guide video generation, thereby unifying saliency anchoring, cache scheduling, and cache management in a principled manner.
    Specifically, \textbf{at the temporal-semantic level}, EchoCache models the time-frequency energy distribution of audio signals as \textit{Cross-modal Saliency Anchors} in the interaction process, and uses it to characterize the potential driving strength of different audio clips for video generation.
    Based on these saliency anchors, \textbf{at the computation-storage level}, we design an \textit{Energy-Guided Step-Latent Cache Scheduling} mechanism, which prioritizes updating critical clips at the latent level while reusing historical cache for non-critical ones, and further adjusts the cache ratio dynamically at the step level according to the state difference, enabling fine-grained quality-efficiency co-optimization.
    Furthermore, since fine-grained caching introduces additional overhead for intermediate-state storage, we propose a \textit{Memory-efficient Cache Management} strategy, which effectively reduces the memory burden of caching through the collaborative storage of reference caches and quantized residuals.

We conduct extensive experiments on mainstream audio-driven video generation models, including Wan2.2-S2V and LongCat-Avatar.
The results show that EchoCache significantly reduces A2V inference latency while preserving generation quality and audio-visual consistency, and achieves a better latency-quality trade-off than existing caching baselines.
For example, on the Wan2.2-S2V model with EMTD benchmark, EchoCache achieves a $2.46\times$ speedup while obtaining the best performance in terms of FID, FVD, and Sync-D, validating the effectiveness of the proposed framework.

% Our main contributions are summarized as follows:
% \begin{itemize}[leftmargin=1.5em]
%     \item \textbf{Problem Analysis.}
%     From the perspective of A2V as a cross-modal upscaling generation process, we reveal two fundamental misalignments of existing caching methods, namely \emph{temporal-semantic misalignment} and \emph{computation-storage misalignment}.

%     \item \textbf{Method Design.}
%     We propose \textbf{EchoCache}, an energy-guided cross-modal caching framework for efficient A2V generation.
%     The framework models the fine-grained temporal dynamics in audio as cross-modal alignment signals for video generation, and unifies latent-level saliency anchoring, cache scheduling, and cache management.

%     \item \textbf{Experimental Results.}
%     We validate the effectiveness of EchoCache on multiple mainstream A2V models and benchmarks.
%     The results demonstrate that EchoCache significantly improves inference efficiency while preserving generation quality and audio-visual consistency, and consistently outperforms existing cache acceleration methods.
% \end{itemize}
\section{Related Work}

\subsection{Audio-driven Video Generation}
Diffusion Transformer (DiT)-based video generation models~\cite{vaswani2017attention,peebles2023scalable} have recently become a dominant paradigm due to their strong generation quality and spatiotemporal modeling capabilities~\cite{yang2024cogvideox,brooks2024video,gao2025seedance,seedance2026seedance2}. Alternative architectures, such as Mamba-attention and multi-modal Mamba, have also been explored for efficient video modeling~\cite{gao2024matten,huang2026m4v}. Early Audio-driven Video (A2V) methods typically adopt a two-stage framework, first mapping audio to intermediate motion representations such as 3DMM or FLAME and then generating videos through GAN-based models or rendering modules~\cite{guan2023stylesync,zhang2023sadtalker,cheng2022videoretalking}. However, their limited intermediate representations and cascading errors make it difficult to jointly ensure realism, temporal consistency, and audio-visual synchronization. Recent methods have shifted toward end-to-end diffusion modeling that jointly learns audio conditions and video dynamics, substantially improving generation quality and synchronization~\cite{gao2025wan,chen2025hunyuanvideo,meituanlongcatteam2025longcatvideoavatartechnicalreport,chen2025seedance15pro}.
Related studies have further extended A2V generation from facial animation to body motion~\cite{lin2025cyberhost,tian2025emo2}, human-object interaction~\cite{zhou2025evaltalker,huang2024magicfight}, and multi-identity or multi-speaker scenarios~\cite{wei2025mocha,huang2025bind}, demonstrating broad potential in digital humans, virtual anchors, and cinematic content creation. Nevertheless, existing methods remain constrained by iterative diffusion sampling, which repeatedly denoises the entire video latent and incurs substantial inference latency, particularly for long videos, thereby limiting practical deployment.

\subsection{Video Generation Acceleration}
Although video generation techniques~\cite{zheng2024open,wan2025wan,team2025longcat} have developed rapidly, the self-attention mechanism requires
 computational complexity of $O(n ^ 2)$, leading to a high computational cost and a long inference time, which severely restricts practical deployment under limited computational resources.
To address this issue, existing acceleration methods mainly focus on model distillation~\cite{luo2023latent,salimans2022progressive,song2023consistency}, quantization compression~\cite{he2023ptqd,shang2023post,feng2025s}, and sparse attention mechanisms ~\cite{xi2025sparse,yuan2024ditfastattn,luo2026training,luo2026attention}.
While these approaches can effectively improve inference efficiency, they typically require additional training or structural modifications to the original model, thereby limiting their applicability to large-scale video generation systems.

Feature cache reuse has recently emerged as a promising direction for accelerating diffusion-based video generation due to its plug-and-play nature and the fact that it does not require retraining.  These methods reduce redundant computation by reusing intermediate features across adjacent diffusion timesteps.
Early caching approaches~\cite{li2023faster,ma2024deepcache,wimbauer2024cache,chen2026ecovideo} primarily adopt static scheduling strategies, where features are reused at fixed intervals along the generation trajectory to achieve stable speedup. 
For example, PAB ~\cite{zhao2024real} selects fixed timestep intervals for each attention block to perform acceleration.
More recent methods introduce dynamic scheduling mechanisms ~\cite{kahatapitiya2025adaptive,lv2024fastercache,liu2025speca} that adaptively determine whether to reuse cached features based on the difference between current and cached representations, thereby reducing cache-induced errors. 
For instance, TeaCache ~\cite{liu2025timestep} and MagCache~\cite{ma2025magcache} estimate output variations by measuring the distance between input features, while TaylorSeer~\cite{liu2025reusing} predicts subsequent features through Taylor expansion, achieving cache-based acceleration from reuse to prediction.

Compared with the above methods, this paper focuses on the Audio-driven Video generation task, introduces audio energy as a saliency anchor for cache scheduling, and further refines cache reuse to the latent granularity, achieving dynamic allocation of computation to critical clips.

\begin{figure*}
  \centering
  \includegraphics[width=\linewidth]{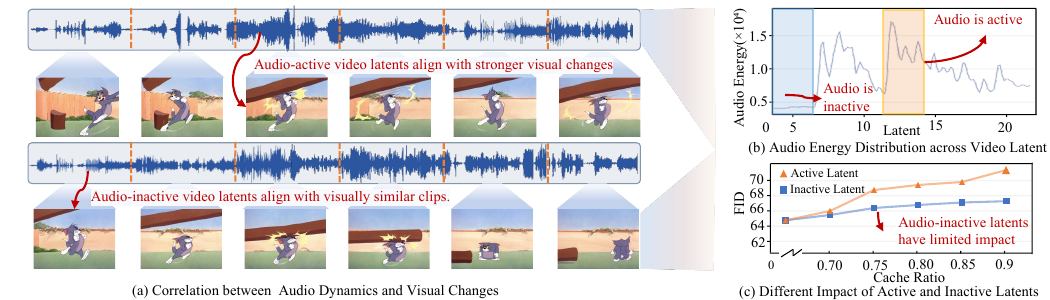}
\caption{(a) A2V is inherently a cross-modal alignment process, where audio dynamics drive corresponding local visual changes. (b) Audio energy distribution in the time-frequency domain provides saliency cues for cross-modal cache scheduling. (c) Impact of cache reuse rates on generation quality for active and inactive latents.}
    \vspace{-6pt}
  \label{fig:3}
\end{figure*}

\section{Methodology}

\subsection{Preliminary}
\label{sec:screenwriter}
\vspace{2pt} 
\noindent \textbf{Audio-driven Video Generation.}
A2V aims to synthesize temporally coherent and audio-visually aligned videos conditioned on an input audio signal.
Most existing methods model this task based on a latent diffusion framework, where a clean video is first mapped into a latent representation $x_0$ via a variational encoder.
For video generation, the latent representation is typically denoted as $x_0 \in \mathbb{R}^{H \times W \times F \times C}$, where $H$, $W$, $F$, and $C$ represent the height, width, number of frames, and number of channels of the latent feature, respectively.
At timestep $t \in \{0, \ldots, T\}$, the forward noising process is defined as
$x_t = a_t x_0 + b_t \epsilon, \epsilon \sim \mathcal{N}(0, I)$,
where $T$ denotes the total number of denoising steps, and $a_t$ and $b_t$ are determined by the noise schedule.
A denoising network conditioned on the audio signal $y$, denoted as $f_\theta(x_t, t, y)$, progressively predicts the clean latent representation through iterative denoising.
    Since the latent evolution is typically smooth both across denoising steps and within each denoising step, there exists significant temporal redundancy during sampling, which provides the foundation for cache-based acceleration.

\vspace{2pt} 
\noindent \textbf{Timestep-level Dynamic Caching.}
In diffusion-based A2V models, the latent representation is encoded as a sequence of spatiotemporal latent tokens.
The intermediate latent output at timestep $t$ is denoted as
$h_t = \{h_t^{(i)}\}_{i=1}^{T \times N}$,
where $h_t^{(i)}$ represents the feature representation of the $i$-th latent token, $T$ denotes the number of latent segments along the temporal dimension, and $N$ denotes the number of spatial latents corresponding to each temporal position.

Existing timestep-level dynamic caching methods~\cite{liu2025timestep,ma2025magcache} mainly exploit the strong redundancy between adjacent denoising steps by reusing intermediate latent features.
Specifically, the relative change between two adjacent timesteps is defined as
\begin{equation}
\Delta_t = \frac{\|h_t - h_{t+1}\|_1}{\|h_{t+1}\|_1 + \xi},
\label{eq:delta}
\end{equation}
where $\xi$ is a small constant for numerical stability.

Assume that the model performs a full computation at timestep $t_a$ and obtains the corresponding intermediate latent representation
$h_{t_a} = \{h_{t_a}^{(i)}\}_{i=1}^{T \times N}$.
This representation is then cached and reused in subsequent timesteps to reduce redundant computation.
Such cache reuse is maintained until the accumulated inter-step feature variation exceeds a predefined threshold $\delta$, beyond which the cached feature is no longer a reliable approximation of the current latent state.
Specifically, let $t_b > t_a$ be the earliest timestep satisfying:
\begin{equation}
\sum_{s=t_a}^{t_b - 1} \Delta_s \le \delta < \sum_{s=t_a}^{t_b} \Delta_s.
\label{eq:cache_trigger}
\end{equation}
For all $t \in [t_a, t_b - 1]$, the current latent feature can be approximated by the cached result as $h_t = h_{t_a}$, and a refresh recomputation is triggered at timestep $t_b$.
    However, existing timestep-level strategies rely mainly on visual similarity, overlooking the cross-modal interaction between audio dynamics and visual synthesis. To address this, EchoCache introduces an energy-guided cross-modal caching framework with latent-level cache scheduling.

\subsection{Analysis}
\label{sec:motivation}
\begin{figure*}
  \centering
  \includegraphics[width=\linewidth]{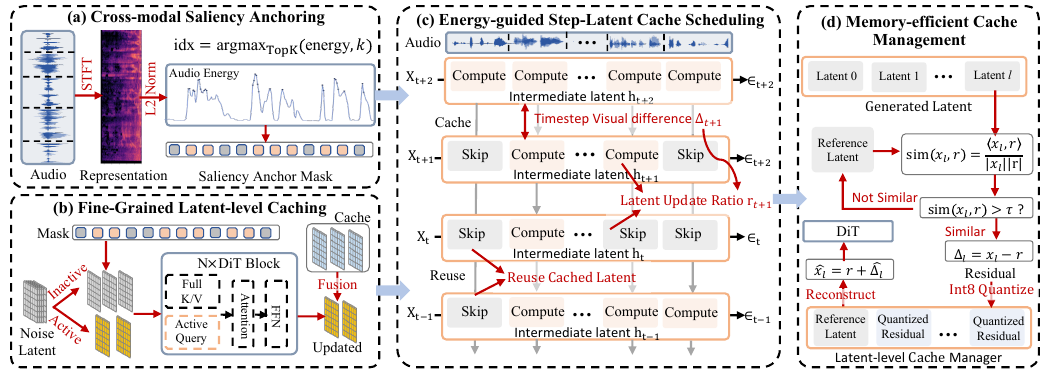}
\caption{Overview of EchoCache. (a) Cross-modal saliency anchors from audio energy. (b) Fine-grained latent-level caching. (c) Energy-guided step-latent cache scheduling. (d) Memory-efficient cache management with adaptive quantization.}
  \label{fig:4}
  \vspace{-3mm}
\end{figure*}

Existing video diffusion caching methods are primarily designed around visual temporal similarity, making cache reuse decisions at the timestep level based on adjacent-step feature differences. However, this design is suboptimal for A2V, where generation is inherently a cross-modal alignment process between audio signals and visual synthesis. As shown in Fig.~\ref{fig:4}(a), audio dynamics affect video latents in a highly non-uniform manner, leading to diverse update demands across temporal clips. Therefore, we analyze latent sensitivity under different levels of audio variation to motivate subsequent fine-grained computation allocation.

\vspace{3pt}
\noindent \textbf{Cross-modal Saliency Analysis.} According to cross-modal saliency cues, we divide video latents into active and inactive groups and analyze their quality sensitivity under different cache ratios. As shown in Fig.~\ref{fig:4}(c), increasing the cache ratio causes a much larger FID increase for active latents than for inactive ones, indicating that audio-active regions are more sensitive to feature reuse and require more frequent updates during denoising. In contrast, inactive latents remain robust even under aggressive caching, with limited quality degradation, suggesting that audio-inactive regions contribute less to final visual fidelity and are more amenable to cache reuse. These results reveal a clear sensitivity gap between latent groups and provide direct support for our fine-grained cache scheduling and computation allocation design.

\vspace{3pt}
\noindent \textbf{Quality Sensitivity Analysis.}
According to cross-modal saliency cues, we divide video latents into active and inactive groups and study their quality sensitivity under different cache ratios. As shown in Fig.~\ref{fig:4}(c), increasing the cache ratio causes a much larger FID increase for active latents than for inactive ones, indicating that audio-active regions are more sensitive to reuse and require more frequent updates. In contrast, inactive latents remain relatively robust even under aggressive caching, suggesting that they have limited impact on final generation quality and are better suited for cache reuse. This observation directly motivates our fine-grained cache scheduling and management design.

\noindent \textbf{Insight for EchoCache.}
This observation directly motivates our cache design philosophy: for A2V, the true potential for cross-step cache reuse lies not in all latents within each denoising step, but in a stable latent subset corresponding to low-change audio segments. Otherwise, unified coarse-grained timestep-level strategies inevitably introduce temporal-semantic misalignment by coupling critical dynamic segments with smooth ones, and further lead to computation-storage misalignment by failing to align computation and cache allocation with audio-driven saliency. Therefore, more effective A2V cache scheduling should move beyond timestep-level reuse and leverage audio-conditioned saliency priors to identify, at the latent level, which tokens are most worthy of updates.

\vspace{-3mm}
\subsection{Energy-Guided Cross-modal Caching}
\label{sec:caching_mechanism}

\noindent \textbf{Cross-modal Saliency Anchors.}
To fully leverage the temporal priors provided by audio conditioning, EchoCache constructs audio energy in the time-frequency domain and establishes a saliency mapping from the audio domain to the latent level.
We term this audio-derived prior the \emph{Cross-modal Saliency Anchor}, as it explicitly transforms fine-grained temporal dynamics in the audio modality into cross-modal alignment signals for video generation.
Specifically, given input audio $S$, we first divide it evenly into $T$ temporal segments and obtain the corresponding time-frequency representation through STFT.
Subsequently, we compute the L2 norm of the time-frequency matrix for each segment to quantify the overall energy response of that audio segment, constructing an audio energy vector
$\text{energy} \in \mathbb{R}^T$.

The analysis in Sec.~\ref{sec:motivation} reveals that the energy distribution of the audio signal in the time-frequency domain closely correlates with the amount of generative information it contains.
Generally, high-energy audio segments correspond to regions with more dramatic signal changes, containing richer time-frequency structure information and thus should not be excessively reused during latent space generation.
Based on this observation, we further map the energy distribution in the audio domain to saliency at the latent level.
Specifically, before generation begins, we employ a Top-K strategy to select indices of high-energy audio segments from the audio energy vector:
$
\text{idx} = \text{argmax}_{\text{topk}}(\text{energy}, k).
$
Here, $\text{idx}$ represents the index set corresponding to high-energy segments, used to locate latent segments more critical for generation.
Based on this, we further construct a latent-level saliency mask with shape $[B, T, N, 1]$ to mark important latents requiring full updates.
To adapt to the dimension organization within self-attention modules, we further flatten this mask to shape $[B, T \times N, 1]$ for direct use during subsequent cache computation.

\vspace{2pt}
\noindent \textbf{Fine-grained latent-level caching.}
With the audio-guided latent-level saliency mask, EchoCache performs latent-level cache computation within each denoising timestep to reduce redundant update overhead for non-critical latents.
This design directly alleviates temporal-semantic misalignment by prioritizing updates on audio-salient latent clips while reusing cached representations for non-salient ones.
Specifically, for the current layer input
$x \in \mathbb{R}^{B \times (T\times N) \times D}$,
the saliency mask constructed from audio energy is
$m \in \{0,1\}^{B \times (T\times N) \times 1}$.
Here, $m_i = 1$ indicates the $i$-th latent belongs to active latents requiring update in the current step, while $m_i = 0$ indicates that latent token can directly reuse the cached result.
Based on this mask, we select active latents from the complete latent sequence and perform current-step main computation only for these positions, thus avoiding redundant updates on all latents.

To improve computational efficiency during fine-grained caching, we extract active latents before attention computation based on the mask and execute query branch computation only on these, while keys and values are still provided by the complete latent set.
Denoting active queries, keys, and values as $Q'$, $K$, and $V$ respectively, attention is computed only on $(Q', K, V)$ to obtain updated results for salient latents.
Subsequently, we restore updated active latents to their original positions in the complete latent layout and fuse with historical cache as:
$
x_{\text{attn}} = m \odot x_{\text{new}} + (1-m) \odot \tilde{x},
$
where $\tilde{x}$ represents cached features and $\odot$ denotes multiplication.

We further extend the same strategy to FFN modules, executing FFN computation only on active latents with other positions directly reusing cached results.
Considering that non-critical latents may experience feature degradation during residual propagation if they continuously skip updates within a block, we perform feature fusion again based on the mask after the FFN residual connection to maintain stability of representations at unupdated positions.

\vspace{2pt}
\noindent \textbf{Energy-Guided Step-Latent Cache Scheduling.}
Although fine-grained latent-level caching reduces redundant updates of non-critical latents within a single denoising step, continuous cache reuse across multiple timesteps still introduces accumulated errors, affecting generation quality.
To address this, EchoCache further introduces a timestep-latent dynamic caching schedule that dynamically adjusts the latent-level cache rate at timestep granularity according to the current generation state, balancing inference efficiency and generation quality.
This design further mitigates computation-storage misalignment by jointly allocating computation and cache reuse across both timestep and latent granularities.

Specifically, considering that early denoising stages exhibit larger latent representation changes and are more sensitive to errors, we retain full computation in early steps and enable fine-grained latent-level caching only in subsequent steps.
Furthermore, to avoid error accumulation from fixed latent cache ratios, we dynamically adjust the current timestep's update rate based on the step-level visual difference indicator in Eq.~\eqref{eq:delta}.
Denoting the normalized result as $\hat{\Delta}_t$, the update rate for the current step is defined as:
$
r_t = \text{clip}\left(r_{\min} + \alpha \hat{\Delta}_t, r_{\min}, r_{\max}\right),
$
where $r_{\min}$ and $r_{\max}$ denote minimum and maximum update ratios respectively, and $\alpha$ is an adjustment coefficient.
Subsequently, for high-saliency latent segments selected by audio energy in the current timestep, we determine the number of active latents according to the update rate: $k_t = \lceil N \cdot r_t \rceil,$ where $N$ denotes the number of latents at each temporal position.

In this manner, when the visual difference in the current timestep is large, $r_t$ increases, preserving more active latents for updates to suppress error accumulation.
When the visual difference is small, $r_t$ decreases, updating only a small number of critical latents while remaining positions directly reuse historical cache results.
Through this approach, EchoCache adaptively allocates computational overhead at both timestep and latent granularities, achieving superior efficiency-quality trade-offs.

\begin{table*}[t]
    \centering
    \caption{Quantitative comparison of EchoCache with baselines on Wan2.2-S2V and LongCat-Avatar. Best and second-best results among acceleration methods are shown in bold and \underline{underlined}. Latency and FLOPs are measured under the same settings.}
    \label{tab:main_quantitative}
    \setlength{\tabcolsep}{7pt}
    \begin{tabular}{l l | c c c c c | c c c c}
        \toprule
        \multirow{2}{*}{\textbf{Dataset}} & \multirow{2}{*}{\textbf{Method}} 
        & \multicolumn{5}{c|}{\textbf{Quality}} 
        & \multicolumn{4}{c}{\textbf{Efficiency}} \\
        \cmidrule(lr){3-7} \cmidrule(lr){8-11}
        & & FID$\downarrow$ & FVD$\downarrow$ & Sync-C$\uparrow$ & Sync-D$\downarrow$ & CSIM$\uparrow$ 
        & Latency(s)$\downarrow$ & Speed$\uparrow$ & PFLOPs$\downarrow$ & Speed$\uparrow$ \\
        \midrule

        \multirow{12}{*}{\textbf{HDTF}}
        & Wan2.2-S2V   & 60.76 & 108.18 & 6.23 & 5.58 & 0.903 & 1039 & 1.00$\times$ & 108.90 & 1.00$\times$ \\
        & TeaCache     & 77.62 & 151.09 & \textbf{6.10} & \underline{6.63} & \underline{0.882} & \underline{542} & \underline{1.92}$\times$ & \underline{59.82} & \underline{1.82}$\times$ \\
        & TaylorSeer   & 76.73 & 135.22 & 5.78 & 7.03 & \underline{0.882} & 803 & 1.29$\times$ & 95.32 & 1.14$\times$ \\
        & MagCache     & \underline{66.91} & \underline{128.71} & \underline{5.94} & 7.00 & \underline{0.882} & 611 & 1.70$\times$ & 63.99 & 1.70$\times$ \\
        & \cellcolor{gray!20}\textbf{EchoCache} & \cellcolor{gray!20}\textbf{66.88} & \cellcolor{gray!20}\textbf{117.67} & \cellcolor{gray!20}\underline{5.94} & \cellcolor{gray!20}\textbf{5.93} & \cellcolor{gray!20}\textbf{0.899} & \cellcolor{gray!20}\textbf{423} & \cellcolor{gray!20}\textbf{2.46}$\times$ & \cellcolor{gray!20}\textbf{54.81} & \cellcolor{gray!20}\textbf{1.98}$\times$ \\
        \cmidrule(lr){2-11}

        & LongCat-Avatar & 51.63 & 206.46 & 9.23 & 6.51 & 0.754 & 742 & 1.00$\times$ & 97.46 & 1.00$\times$ \\
        & TeaCache       & 61.88 & 311.04 & \underline{8.88} & \underline{7.83} & 0.654 & \underline{467} & \underline{1.58}$\times$ & \underline{61.63} & \underline{1.58}$\times$ \\
        & TaylorSeer     & 81.43 & 282.47 & 8.44 & 7.89 & 0.667 & 535 & 1.39$\times$ & 68.15 & 1.43$\times$ \\
        & MagCache       & \underline{61.08} & \underline{254.52} & 8.50 & 7.89 & \underline{0.668} & 510 & 1.45$\times$ & 65.85 & 1.48$\times$ \\
        & \cellcolor{gray!20}\textbf{EchoCache} & \cellcolor{gray!20}\textbf{57.82} & \cellcolor{gray!20}\textbf{215.75} & \cellcolor{gray!20}\textbf{9.08} & \cellcolor{gray!20}\textbf{6.83} & \cellcolor{gray!20}\textbf{0.702} & \cellcolor{gray!20}\textbf{413} & \cellcolor{gray!20}\textbf{1.80}$\times$ & \cellcolor{gray!20}\textbf{52.26} & \cellcolor{gray!20}\textbf{1.86}$\times$ \\

        \midrule

        \multirow{12}{*}{\textbf{EMTD}}
        & Wan2.2-S2V   & 65.66 & 129.57 & 6.51 & 5.95 & 0.877 & 1039 & 1.00$\times$ & 108.94 & 1.00$\times$ \\
        & TeaCache     & 100.23 & \underline{174.14} & 5.65 & 6.83 & 0.799 & \underline{542} & \underline{1.92}$\times$ & \underline{59.82} & \underline{1.82}$\times$ \\
        & TaylorSeer   & \underline{79.58} & 176.15 & 6.26 & 6.80 & 0.810 & 803 & 1.29$\times$ & 95.32 & 1.14$\times$ \\
        & MagCache     & 110.78 & 199.00 & \textbf{6.49} & \underline{6.72} & \textbf{0.817} & 611 & 1.70$\times$ & 63.99 & 1.70$\times$ \\
        & \cellcolor{gray!20}\textbf{EchoCache} & \cellcolor{gray!20}\textbf{76.01} & \cellcolor{gray!20}\textbf{162.66} & \cellcolor{gray!20}\underline{6.39} & \cellcolor{gray!20}\textbf{6.01} & \cellcolor{gray!20}\underline{0.816} & \cellcolor{gray!20}\textbf{423} & \cellcolor{gray!20}\textbf{2.46}$\times$ & \cellcolor{gray!20}\textbf{54.81} & \cellcolor{gray!20}\textbf{1.98}$\times$ \\
        \cmidrule(lr){2-11}

        & LongCat-Avatar & 65.05 & 433.26 & 8.71 & 6.93 & 0.672 & 742 & 1.00$\times$ & 97.46 & 1.00$\times$ \\
        & TeaCache       & 113.15 & 527.19 & 7.36 & 8.19 & \underline{0.615} & \underline{467} & \underline{1.58}$\times$ & \underline{61.63} & \underline{1.58}$\times$ \\
        & TaylorSeer     & \underline{76.74} & 652.68 & 7.52 & \underline{8.03} & 0.602 & 535 & 1.39$\times$ & 68.15 & 1.43$\times$ \\
        & MagCache       & 114.33 & \underline{492.99} & \underline{7.55} & 8.06 & 0.612 & 510 & 1.45$\times$ & 65.85 & 1.48$\times$ \\
        & \cellcolor{gray!20}\textbf{EchoCache} & \cellcolor{gray!20}\textbf{74.85} & \cellcolor{gray!20}\textbf{467.12} & \cellcolor{gray!20}\textbf{8.48} & \cellcolor{gray!20}\textbf{7.07} & \cellcolor{gray!20}\textbf{0.621} & \cellcolor{gray!20}\textbf{412} & \cellcolor{gray!20}\textbf{1.80}$\times$ & \cellcolor{gray!20}\textbf{52.26} & \cellcolor{gray!20}\textbf{1.86}$\times$ \\

        \bottomrule
    \end{tabular}
\end{table*}

\subsection{Memory-efficient Cache Management}
\label{sec:cache_manager}

Although fine-grained timestep-latent caching effectively reduces redundant computation, it typically requires maintaining a separate cache for multiple latents in each denoising step, causing cache consumption to grow linearly with the number of modules.
To resolve the additional storage burden introduced by fine-grained reuse, we further design a Memory-efficient Cache Management strategy with adaptive quantization, which improves the alignment between computation scheduling and storage management.

We observe that cached representations of adjacent latents within the same denoising step typically exhibit high similarity.
Therefore, rather than always storing the complete cache output $x_l$ for the $l$-th latent in the current timestep, we first compute its cosine similarity with the current reference cache $r$:
$
\text{sim}(x_l, r) = \frac{\langle x_l, r \rangle}{\|x_l\|_2 \|r\|_2}.
$
When $\text{sim}(x_l, r) > \tau$, where $\tau$ is the cache quantization threshold, we store only the residual relative to the reference cache $\Delta_l = x_l - r$ and quantize $\Delta_l$ to int8.
Otherwise, we store $x_l$ as the new reference cache in full precision.
Through this design, the cache manager adaptively switches between full cache and reference-cache-plus-quantized-residual forms to reduce storage redundancy.

During cache retrieval, if the target latent is stored in residual form, we reconstruct its output according to reference cache $r$ and dequantized residual $\hat{\Delta}_l$:
$
\hat{x}_l = r + \hat{\Delta}_l.
$
The reconstructed result is then converted back to the original computation precision to ensure consistency for subsequent operations.

Additionally, since reference caches evolve over time, we introduce a reference-based release mechanism: before writing new cache, we first delete old residual entries and check whether their dependent reference caches are still referenced by other latents.
If not, we release them as well.
Through this design, EchoCache compresses the additional GPU memory overhead introduced by fine-grained caching without affecting generation quality.

\section{Experiments}

\subsection{Experimental Setup}

\begin{figure*}
  \centering
  \includegraphics[width=\linewidth]{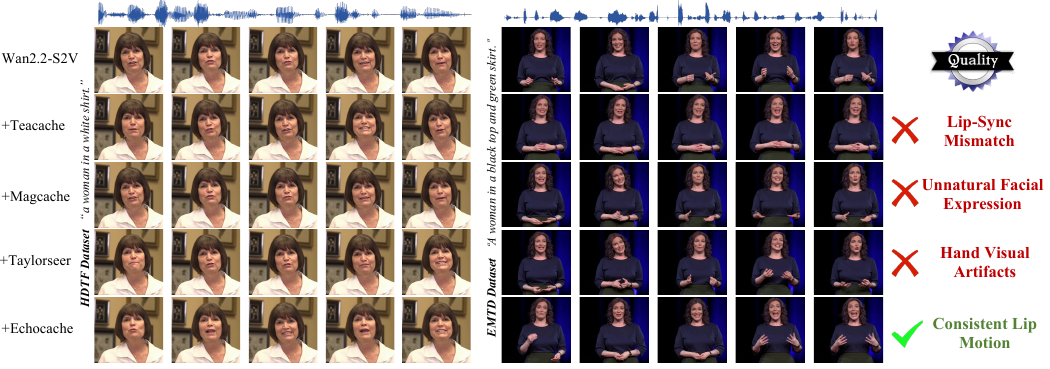}
\caption{Qualitative comparison of EchoCache and baseline methods on Wan2.2-S2V across the HDTF and EMTD datasets.}
  \label{fig:5}
\end{figure*}

\noindent\textbf{Models and Baselines.}
We evaluate the proposed method on two open-source audio-driven video (A2V) generation models, namely Wan2.2-S2V~\cite{gao2025wan} and LongCat-Avatar~\cite{meituanlongcatteam2025longcatvideoavatartechnicalreport}. We further compare our method with three SOTA caching-based acceleration methods, including TeaCache~\cite{liu2025timestep}, MagCache~\cite{ma2025magcache}, and TaylorSeer~\cite{liu2025reusing}. To ensure a fair comparison, all methods are evaluated under strictly aligned hyperparameters in the video generation process. Specifically, Wan2.2-S2V is used to generate 5-second videos with 81 frames at 720P resolution, while LongCat-Avatar is used to generate 5-second videos with 93 frames at 480P resolution. All experiments are conducted on NVIDIA H200 141GB GPUs.

\noindent\textbf{Datasets.}
To verify the effectiveness of the proposed method, we follow the experimental settings of Wan2.2-S2V and LongCatAvatar~\cite{gao2025wan,meituanlongcatteam2025longcatvideoavatartechnicalreport}, and adopt two types of datasets for evaluation: talking-head and talking-body. Specifically, HDTF~\cite{zhang2021flow} is used for talking-head evaluation, and EMTD~\cite{meng2024echomimicv2} is used for talking-body evaluation.

\noindent\textbf{Evaluation Metrics.}
We evaluate the proposed method from two aspects: quality and efficiency. For quality, we adopt FID~\cite{heusel2017gans}, FVD~\cite{unterthiner2019fvd}, Sync-C, Sync-D~\cite{chung2016out}, and CSIM to comprehensively measure the visual quality, audio-visual synchronization, and identity consistency of the generated videos. For efficiency, we report Latency, FLOPs, and Speedup.

\begin{figure*}
  \centering
  \includegraphics[width=\linewidth]{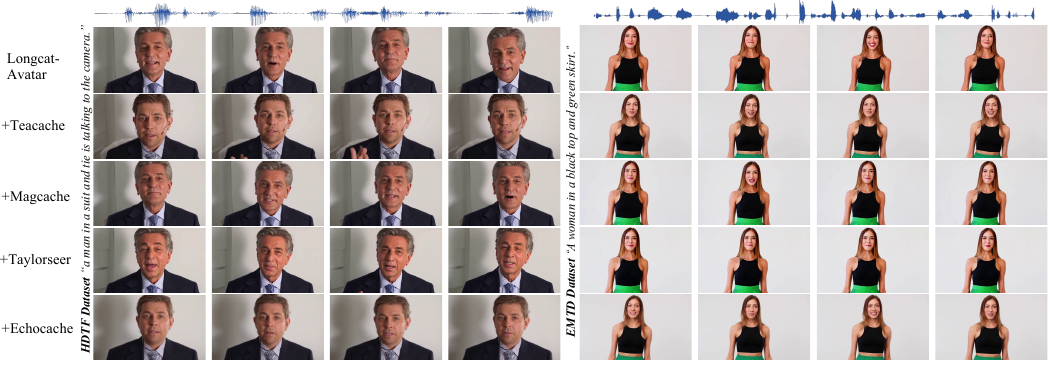}
\caption{Qualitative comparison of EchoCache and baselines on LongCat-Avatar across the HDTF and EMTD datasets. }
  \label{fig:6}
\end{figure*}

\subsection{Main Results}

\noindent\textbf{Quantitative Comparison on HDTF.}
We evaluate EchoCache on HDTF to measure the quality--efficiency trade-off in talking-head generation, as shown in Table \ref{tab:main_quantitative}. On Wan2.2-S2V, EchoCache reduces latency from 1039 s to 423 s and FLOPs from 108.9 to 54.81, achieving 2.46$\times$ and 1.98$\times$ speedups, respectively. It also improves generation quality over caching baselines, reducing FID and FVD to 66.88 and 117.67, while achieving a CSIM of 0.899 and maintaining performance close to the original model. These results show that EchoCache substantially lowers inference cost while preserving visual details, identity consistency, and lip synchronization.

\noindent\textbf{Quantitative Comparison on EMTD.}
We further evaluate EchoCache on EMTD in the more challenging talking-body scenario. On LongCat-Avatar, it reduces latency from 742 s to 412 s and FLOPs from 97.46 to 52.26, corresponding to 1.80$\times$ and 1.86$\times$ speedups. EchoCache achieves the lowest FVD of 467.12 among caching methods, together with a Sync-C of 8.48, a Sync-D of 7.07, and a CSIM of 0.621, while closely matching the original model. These results demonstrate a favorable quality--efficiency trade-off under complex body motion and spatiotemporal variation, confirming its robust cross-scenario generalization across diverse motion patterns.

\noindent\textbf{Qualitative Visualizations of Different Methods.}
The qualitative comparisons in Fig.~\ref{fig:5} and Fig.~\ref{fig:6} show that existing caching methods often introduce lip-sync errors, detail blurring, and temporal inconsistency at high acceleration ratios, especially under complex articulation and motion. In contrast, EchoCache better preserves appearance, fine-grained lip dynamics, and temporal coherence across HDTF and EMTD on both Wan2.2-S2V and LongCat-Avatar. These results further demonstrate its superior quality--efficiency trade-off over prior caching baselines.

\vspace{-2mm}
\subsection{Ablation Study}

\begin{table}[t]
    \centering
    \caption{Module ablation of EchoCache on Wan2.2-S2V.}
    \label{tab:ablation_components}
    \setlength{\tabcolsep}{4pt}
    \vspace{-6pt}
    \resizebox{\columnwidth}{!}{
    \begin{tabular}{lccccc}
        \toprule
        \textbf{Method} & FVD$\downarrow$ & Sync-C$\uparrow$ & CSIM$\uparrow$ & Latency(s)$\downarrow$ & Peak Memory(GB)$\downarrow$ \\
        \midrule
        Wan2.2-S2V & 129.57 & 6.51 & 0.877 & 1039 & 57.23 \\
        +TaylorSeer& 176.15 & 6.26 & 0.810 & 803 & 99.16 \\
        +EchoCache & 162.66 & 6.39 & 0.816 & 423 & 59.12 \\
        \midrule
        
        \multicolumn{6}{l}{\textbf{EchoCache without Anchor Selection Strategy}} \\
        Uniform & 176.05 & 6.24 & 0.821 & 412 & 58.99 \\
        Random & 192.26 & 6.28 & 0.825 & 420 & 59.04 \\
        \midrule
        
        \multicolumn{6}{l}{\textbf{EchoCache without Cache Scheduling Strategy}} \\
        Step-only & 174.14 & 5.65 & 0.799 & 542 & 58.34 \\
        Latent-only & 169.21 & 5.72 & 0.801 & 402 & 68.17 \\
        \midrule
        
        \multicolumn{6}{l}{\textbf{EchoCache without Cache Quantization Strategy}} \\
        FP16 & 163.11 & 6.63 & 0.841 & 435 & 81.02 \\
        Naive INT8 & 183.26 & 6.01 & 0.805 & 445 & 58.57 \\
        
        \bottomrule
    \end{tabular}
    }
\end{table}

\begin{figure}
  \centering
  \includegraphics[width=\linewidth]{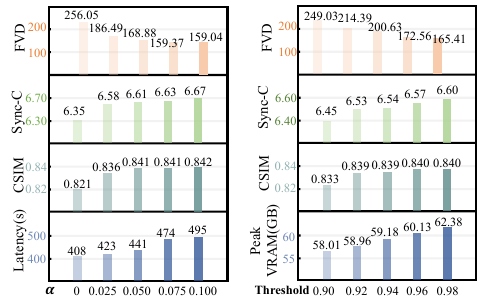}
\caption{Hyperparameter sensitivity analysis of EchoCache, showing the effects of update ratio coefficient $\alpha$ and cache similarity threshold on quality, latency, and peak VRAM.}
  \label{fig:7}
  \vspace{-3mm}
\end{figure}

\noindent\textbf{Impact of the Anchor Selection Strategy.}
To verify the effectiveness of the anchor selection strategy, we replace it with Uniform and Random selection. As shown in Table.~\ref{tab:ablation_components}, removing the audio-energy-based anchor selection leads to clear performance degradation: FVD rises to 176.05 and 192.26, while Sync-C drops to 6.24 and 6.28, respectively. In contrast, full EchoCache achieves a better FVD of 162.66 and Sync-C of 6.39, showing that audio-energy-guided saliency modeling can more accurately identify critical latents.

\noindent\textbf{Impact of the Cache Scheduling Strategy.}
As shown in Table.~\ref{tab:ablation_components}, we further analyze the effect of the cache scheduling strategy. Using only the Step-only strategy increases latency to 542 s and worsens FVD to 174.14. Using only the Latent-only strategy reduces latency to 402 s, but significantly increases peak memory to 68.17 GB, while FVD remains 169.21. By comparison, full EchoCache achieves a better FVD of 162.66 with 423s latency and 59.12 GB peak memory. These results show that step-latent collaborative scheduling is necessary for balancing quality, efficiency, and memory overhead.

\noindent\textbf{Impact of the Cache Quantization Strategy.}
As shown in Table.~\ref{tab:ablation_components}, we evaluate the effect of the cache quantization strategy on quality and memory usage. Direct FP16 storage achieves relatively good Sync-C of 6.63 and CSIM of 0.841, but increases peak memory to 81.02 GB. In contrast, Naive INT8 reduces memory to 58.57 GB, but degrades FVD to 183.26, indicating large reconstruction errors. Full EchoCache, however, maintains an FVD of 162.66 and a Sync-C of 6.39 with only 59.12 GB peak memory, demonstrating that the proposed cache quantization strategy effectively reduces memory overhead while alleviating quality loss caused by quantization.

\noindent\textbf{Hyperparameter Sensitivity Analysis.}
As shown in Fig.~\ref{fig:7}, we analyze the sensitivity of EchoCache to $\alpha$ and $\tau$. Increasing $\alpha$ improves quality, but incurs higher latency, revealing the quality–efficiency trade-off of dynamic scheduling. Increasing $\tau$ similarly improves generation quality, while raising peak memory due to less aggressive cache quantization. Overall, $\alpha$ controls the quality–latency trade-off, whereas $\tau$ determines the quality–memory trade-off.

\begin{figure}
  \centering
  \includegraphics[width=\linewidth]{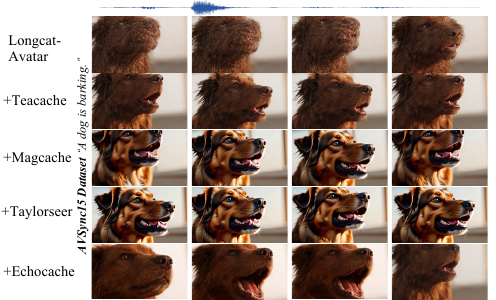}
  \vspace{-12pt}
\caption{Qualitative ablation results on AVSync15. Compared with baseline caching methods, EchoCache better preserves appearance details and motion coherence in motion-rich audio-driven video generation.}
  \label{fig:8}
  \vspace{-6mm}
\end{figure}

\noindent\textbf{Ablation Study on AVSync15.}
As shown in Fig.~\ref{fig:8}, we further conduct ablation experiments on AVSync15~\cite{linz2024asva} to evaluate the generality of EchoCache beyond the talking-head and talking-body settings used in the main experiments. AVSync15 is a subset of VGG-Sound containing fifteen classes of in-the-wild activities with highly synchronized audio and video. The results show that EchoCache remains effective on AVSync15, further confirming the generalizability of the proposed energy-guided caching strategy.

\section{Limitations and Future Directions}

EchoCache still has limitations. Its audio-energy prior may not fully capture richer cross-modal cues, and its cache scheduling remains partly heuristic across models and settings. Future work may extend the framework to broader multimodal conditions, such as video inputs and other motion-related trajectory modalities, to enable more robust and generalizable modality-aware caching for video generation. We hope this work can inspire unified modality-aware caching frameworks for fine-grained computation allocation.

\section{Conclusion}

In this paper, we identify temporal-semantic and computation-storage misalignments in existing caching methods for audio-driven video generation and propose EchoCache, an energy-guided cross-modal caching framework. By combining cross-modal saliency anchoring, step-latent cache scheduling, and memory-efficient cache management, EchoCache enables fine-grained computation allocation based on audio dynamics. We hope this work advances modality-aware acceleration for diffusion-based video generation.

%=======================================%
%            Acknowledgments            %
%=======================================%

% 如有致谢，必须使用 acks 环境，并放在参考文献之前。
% \begin{acks}
% This work was supported by ...
% \end{acks}

%=======================================%
%               Appendix                %
%=======================================%

%
% \appendix
% \input{_txt/appendix}

%=======================================%
%             Bibliography              %
%=======================================%

\bibliographystyle{ACM-Reference-Format}
\balance
\bibliography{_ref/ref}

\end{document}